\pdfoutput=1

\documentclass{article} 
\usepackage{nips13submit_e,times}
\usepackage{hyperref}
\usepackage{url}
\usepackage{amssymb, amsmath}
\usepackage{epsfig}
\usepackage{array}
\usepackage{ifthen}
\usepackage{color}
\usepackage{fancyhdr}
\usepackage{graphicx}

\title{ Deconvolution of High Dimensional Mixtures via Boosting, 
  with Application to Diffusion-Weighted MRI of Human Brain }

\author{
Charles Y.~Zheng \\
Department of Statistics\\
Stanford University\\
Stanford, CA 94305 \\
\texttt{snarles@stanford.edu} \\
\And
Franco ~Pestilli \\
Department of Psychological and Brain Sciences\\
Indiana University, Bloomington, IN 47405\\
\texttt{franpest@indiana.edu} \\
\AND
Ariel ~Rokem \\
Department of Psychology\\
Stanford University\\
Stanford, CA 94305 \\
\texttt{arokem@stanford.edu} \\
}

\nipsfinalcopy 

\begin{document}

\maketitle

\begin{abstract}
  Diffusion-weighted magnetic resonance imaging (DWI) and fiber tractography
  are the only methods to measure the structure of the white matter in the
  living human brain. The diffusion signal has been modelled as the combined
  contribution from many individual fascicles of nerve fibers passing through
  each location in the white matter. Typically, this is done via \emph{basis
    pursuit}, but estimation of the exact directions is limited due to
  discretization [1, 2]. The difficulties inherent in modeling DWI data are
  shared by many other problems involving fitting non-parametric mixture
  models. Ekanadaham et al. [3] proposed an approach, \emph{continuous basis
    pursuit}, to overcome discretization error in the 1-dimensional case (e.g.,
  spike-sorting). Here, we propose a more general algorithm that fits mixture
  models of any dimensionality without discretization. Our algorithm uses the
  principles of L2-boost [4], together with refitting of the weights and
  pruning of the parameters. The addition of these steps to L2-boost both
  accelerates the algorithm and assures its accuracy.  We refer to the
  resulting algorithm as \emph{elastic basis pursuit}, or EBP, since it expands
  and contracts the active set of kernels as needed. We show that in contrast
  to existing approaches to fitting mixtures, our boosting framework (1)
  enables the selection of the optimal bias-variance tradeoff along the
  solution path, and (2) scales with high-dimensional problems. In simulations
  of DWI, we find that EBP yields better parameter estimates than a non-negative
  least squares (NNLS) approach, or the standard model used in DWI, the tensor
  model, which serves as the basis for diffusion tensor imaging (DTI) [5]. We
  demonstrate the utility of the method in DWI data acquired in parts of the
  brain containing crossings of multiple fascicles of nerve fibers.
\end{abstract}

\section{Introduction}

In many applications, one obtains measurements $(x_i, y_i)$ for which the
response $y$ is related to $x$ via some mixture of known kernel functions
$f_\theta(x)$, and the goal is to recover the mixture parameters $\theta_k$ and
their associated weights: 

\begin{equation}\label{mixtureeq}
y_i = \sum_{k=1}^K w_k f_{\theta_k}(x) + \epsilon_i
\end{equation}

where $f_\theta(x)$ is a known kernel function parameterized by $\theta$, and
$\boldsymbol{\theta} = (\theta_1,\hdots,\theta_K)$ are model parameters to be
estimated, $w= (w_1,\hdots,w_K)$ are unknown nonnegative weights to be
estimated, and $\epsilon_i$ is additive noise. The number of components $K$ is
also unknown, hence, this is a \emph{nonparametric model}. One example of a
domain in which mixture models are useful is the analysis of data from
diffusion-weighted magnetic resonance imaging (DWI). This biomedical imaging
technique is sensitive to the direction of water diffusion within
millimeter-scale voxels in the human brain \emph{in vivo}. Water molecules
freely diffuse along the length of nerve cell axons, but is restricted by cell
membranes and myelin along directions orthogonal to the axon's
trajectory. Thus, DWI provides information about the microstructural properties
of brain tissue in different locations, about the trajectories of organized
bundles of axons, or fascicles within each voxel, and about the connectivity
structure of the brain. Mixture models are employed in DWI to deconvolve the
signal within each voxel with a kernel function, $f_{\theta}$, assumed to
represent the signal from every individual fascicle [1, 2] (Figure
\ref{fig:illus1}B), and $w_i$ provide an estimate of the fiber orientation
distribution function (fODF) in each voxel, the direction and volume fraction
of different fascicles in each voxel. In other applications of mixture modeling
these parameters represent other physical quantities. For example, in
chemometrics, $\theta$ represents a chemical compound and $f_\theta$ its
spectra. In this paper, we focus on the application of mixture models to the
data from DWI experiments and simulations of these experiments.

\subsection{Model fitting - existing approaches}

Hereafter, we restrict our attention to the use of squared-error loss;
resulting in penalized least-squares problem
\begin{equation}\label{lsp}
\text{minimize }_{\hat{K},\hat{w},\hat{\boldsymbol{\theta}}}\left\|y_i -
  \sum_{k=1}^{\hat{K}}  \hat{w}_k f_{\hat{\theta}_k}(x_i)\right\|^2 +
\lambda P_{\boldsymbol{\theta}}(w)
\end{equation}

\begin{figure}[htbp]
\centering
\includegraphics[scale=0.2]{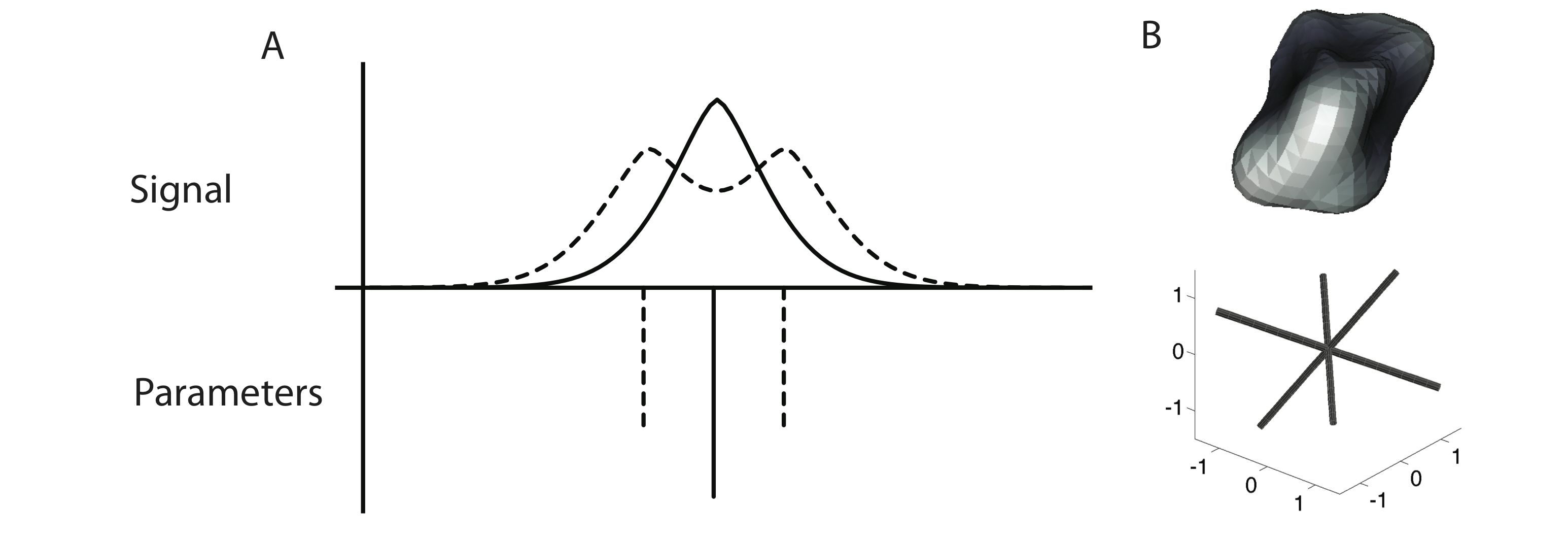}
\caption{The signal deconvolution problem. Fitting a mixture model with a NNLS
  algorithm is prone to errors due to discretization. For example, in 1D (A),
  if the true signal (top; dashed line) arises from a mixture of signals from a
  bell-shaped kernel functions (bottom; dashed line), but only a single kernel
  function between them is present in the basis set (bottom; solid line),
  this may result in inaccurate signal predictions (top; solid line), due to
  erroneous estimates of the parameters $w_i$. This problem arises in
  deconvolving multi-dimensional signals, such as the 3D DWI signal (B), as
  well. Here, the DWI signal in an individual voxel is presented as a 3D
  surface (top). This surface results from a mixture of signals arising from
  the fascicles presented on the bottom passing through this single (simulated)
  voxel. Due to the signal generation process, the kernel of the diffusion
  signal from each one of the fascicles has a minimum at its center, resulting
  in 'dimples' in the diffusion signal in the direction of the peaks in the
  fascicle orientation distribution function.}
\label{fig:illus1}
\end{figure}

Minimization problems of the form \eqref{lsp} can be found in the signal
deconvolution literature and elsewhere: some examples include super-resolution
in imaging [6], entropy estimation for discrete distributions [7], X-ray
diffraction [8], and neural spike sorting [3]. Here,
$P_{\boldsymbol{\theta}}(w)$ is a \emph{convex} penalty function of
$(\boldsymbol{\theta}, w)$.  Examples of such penalty functions given in
Section \ref{s:regularization}; a formal definition of convexity in the
nonparametric setting can be found in the supplementary material, but will not
be required for the results in the paper. Technically speaking, the objective
function \eqref{lsp} is convex in $(w,\boldsymbol{\theta})$, but since its
domain is of infinite dimensionality, for all practical purposes \eqref{lsp} is
a nonconvex optimization problem.  One can consider fixing the number of
components in advance, and using a descent method (with random restarts) to
find the best model of that size.  Alternatively, one could use a stochastic
search method, such as simulated annealing or MCMC [9], to estimate the size of
the model and the model parameters simultaneously.  However, as one begins to
consider fitting models with increasing number of components $\hat{K}$ and of
high dimensionality, it becomes increasingly difficult to apply these
approaches [3]. Hence a common approach to obtaining an approximate solution to
\eqref{lsp} is to limit the search to a discrete grid of candidate parameters
$\boldsymbol{\theta} = \theta_1,\hdots,\theta_p$.  The estimated weights and
parameters are then obtained by solving an optimization problem of the form
\[
\hat{\beta} = \text{argmin}_{\beta >0} ||y-\vec{F}\beta||^2 + \lambda P_{\boldsymbol{\theta}}(\beta)
\]
where $\vec{F}$ has the $j$th column $\vec{f}_{\theta_j}$, where
$\vec{f}_\theta$ is defined by $(\vec{f}_\theta)_i = f_{\theta}(x_i)$.
Examples applications of this non-negative least-squares-based approach (NNLS)
include [10] and [1, 2, 7]. In contrast to descent based methods, which get
trapped in local minima, NNLS is guaranteed to converge to a solution which is
within $\epsilon$ of the global optimum, where $\epsilon$ depends on the scale
of discretization. In some cases, NNLS will predict the signal accurately (with
small error), but the parameters resulting will still be erroneous. Figure
\ref{fig:illus1} illustrates the worst-case scenario where discretization is
misaligned relative to the true parameters/kernels that generated the signal.

In an effort to improve the discretization error of NNLS, Ekanadham et al [3]
introduced continuous basis pursuit (CBP).  CBP is an extension of nonnegative
least squares in which the points on the discretization grid
$\theta_1,\hdots,\theta_p$ can be continuously moved within a small distance;
in this way, one can reach any point in the parameter space.  But instead of
computing the actual kernel functions for the perturbed parameters, CBP uses
linear approximations, e.g. obtained by Taylor expansions. Depending on the
type of approximation employed, CBP may incur large error. The developers of
CBP suggest solutions for this problem in the one-dimensional case, but these
solutions cannot be used for many applications of mixture models (e.g DWI). The
computational cost of both NNLS and CBP scales exponentially in the
dimensionality of the parameter space.  In contrast, using stochastic search
methods or descent methods to find the global minimum will generally incur a
computational cost scaling which is exponential in the sample size times the
parameter space dimensions. Thus, when fitting high-dimensional mixture models,
practitioners are forced to choose between the discretization errors inherent
to NNLS, or the computational difficulties in the descent methods.  We will
show that our boosting approach to mixture models combines the best of both
worlds: while it does not suffer from discretization error, it features
computational tractability comparable to NNLS and CBP. We note that for the
specific problem of super-resolution, C{\`a}ndes derived a deconvolution
algorithm which finds the global minimum of \eqref{lsp} without discretization
error and proved that the algorithm can recover the true parameters under a
minimal separation condition on the parameters [6].  However, we are unaware of
an extension of this approach to more general applications of mixture models.

\subsection{Boosting}

The model \eqref{mixtureeq} appears in an entirely separate context, as the
model for learning a regression function as an ensemble of weak learners
$f_\theta$, or boosting [4].  However, the problem of fitting a mixture model
and the problem of fitting an ensemble of weak learners have several important
differences.  In the case of learning an ensemble, the family $\{f_\theta\}$
can be freely chosen from a universe of possible weak learners, and the only
concern is minimizing the prediction risk on a new observation.  In contrast,
in the case of fitting a mixture model, the family $\{f_\theta\}$ is specified
by the application. As a result, boosting algorithms, which were derived under
the assumption that $\{f_\theta\}$ is a suitably flexible class of weak
learners, generally perform poorly in the signal deconvolution setting, where
the family $\{f_\theta\}$ is inflexible. In the context of regression,
$L_2$boost, proposed by Buhlmann et al [4] produces a path of ensemble models
which progressively minimize the sum of squares of the residual. $L_2$boost
fits a series of models of increasing complexity.  The first model consists of
the single weak learner $\vec{f}_\theta$ which best fits $y$.  The second model
is formed by finding the weak learner with the greatest correlation to the
residual of the first model, and adding the new weak learner to the model,
without changing any of the previously fitted weights.  In this way the size of
the model grows with the number of iterations: each new learner is fully fit to
the residual and added to the model.  But because the previous weights are
never adjusted, $L_2$Boost fails to converge to the global minimum of
\eqref{lsp} in the mixture model setting, producing suboptimal solutions. In
the following section, we modify $L_2$Boost for fitting mixture models.  We
refer to the resulting algorithm as \emph{elastic basis pursuit}.

\section{Elastic Basis Pursuit}\label{s:tcbp}

Our proposed procedure for fitting mixture models consists of two stages.  In
the first stage, we transform a $L_1$ penalized problem to an equivalent
\emph{non regularized} least squares problem.  In the second stage, we employ a
modified version of $L_2$Boost, \emph{elastic basis pursuit}, to solve the
transformed problem.  We will present the two stages of the procedure, then
discuss our fast convergence results.

\subsection{Regularization}\label{s:regularization}

For most mixture problems it is beneficial to apply a $L_1$-norm based penalty,
by using a modified input $\tilde{y}$ and kernel function family
$\tilde{f}_\theta$, so that
\begin{equation}
\label{tildessr}
\text{argmin}_{K, w, \boldsymbol{\theta}}
\left\| y - \sum_{i=1}^K \vec{f}_\theta \right \|^2 + \lambda
P_{\boldsymbol{\theta}}(w)
=\text{argmin}_{K, w,\boldsymbol{\theta}} \left\| \tilde{y} - \sum_{i=1}^K
  \tilde{f}_\theta \right\|^2
\end{equation}

We will use our modified $L_2$Boost algorithm to produce a path of
solutions for objective function on the left side, which results in a
solution path for the penalized objective function \eqref{lsp}.

For example, it is possible to embed the penalty $P_{\boldsymbol{\theta}}(w) =
||w||_1^2$ in the optimization problem \eqref{lsp}.  One can show that
solutions obtained by using the penalty function $P_{\boldsymbol{\theta}}(w) =
||w||_1^2$ have a one-to-one correspondence with solutions of obtained using
the usual $L_1$ penalty $||w||_1$.  The penalty $||w||^2_1$ is implemented by
using the transformed input: $\tilde{y}=\begin{pmatrix}y\\0\end{pmatrix}$ and
using modified kernel vectors $\tilde{f}_\theta
= \begin{pmatrix}\vec{f}_\theta\\\sqrt{\lambda}\end{pmatrix}$. Other kinds of
regularization are also possible, and are presented in the \emph{supplemental
  material}.

\subsection{From $L_2$Boost to Elastic Basis Pursuit}

Motivated by the connection between boosting and mixture modelling, we consider
application of $L_2$Boost to solve the transformed problem (the left side
of\eqref{tildessr}). Again, we reiterate the \emph{nonparametric} nature of the
model space; by minimizing \eqref{tildessr}, we seek to find the model with
\emph{any} number of components which minimizes the residual sum of squares.
In fact, given appropriate regularization, this results in a well-posed
problem. In each iteration of our algorithm a subset of the parameters,
$\boldsymbol{\theta}$ are considered for adjustment. Following Lawson and
Hanson [11], we refer to these as the \emph{active set}. As stated before,
$L_2$Boost can only grow the active set at each iteration, converging to
inaccurate models.  Our solution to this problem is to modify $L_2$Boost so
that it grows \emph{and} contracts the active set as needed; hence we refer to
this modification of the $L_2$Boost algorithm as \emph{elastic basis
  pursuit}. The key ingredient for any boosting algorithm is an oracle for
fitting a weak learner: that is, a function $\tau$ which takes a residual as
input and returns the parameter $\theta$ corresponding to the kernel
$\tilde{f}_\theta$ most correlated with the residual. EBP takes as inputs the
oracle $\tau$, the input vector $\tilde{y}$, the function $\tilde{f}_\theta$,
and produces a path of solutions which progressively minimize \eqref{tildessr}.
To initialize the algorithm, we use NNLS to find an initial estimate of
$(w,\boldsymbol{\theta})$.  In the $k$th iteration of the boosting algorithm,
let $\tilde{r}^{(k-1)}$ be residual from the previous iteration (or the NNLS
fit, if $k=1$).  The algorithm proceeds as follows
\begin{enumerate}
\item
Call the oracle to find $\theta_{new}=\tau(\tilde{r}^{(k-1)})$, and add
$\theta_{new}$ to the active set $\boldsymbol{\theta}$.
\item 
Refit the weights $w$, using NNLS, to solve:
\[
\text{minimize}_{w > 0} ||\tilde{y}-\tilde{F}w||^2
\]
where $\tilde{F}$ is the matrix formed from the regressors in the active set,
$\tilde{f}_\theta$ for $\theta \in \boldsymbol{\theta}$.
This yields the residual $\tilde{r}^{(k)} = \tilde{y}-\tilde{F}w$.
\item
Prune the active set $\boldsymbol{\theta}$ by removing any
parameter $\theta$ whose weight is zero, and update the weight vector
$w$ in the same way.
This ensures that the active set $\boldsymbol{\theta}$ remains sparse
in each iteration.
Let $(w^{(k)},\boldsymbol{\theta}^{(k)})$ denote the values of
$(w,\boldsymbol{\theta})$ at the end of this step of the iteration.
\item Stopping may be assessed by computing an estimated prediction error at
  each iteration, via an independent validation set, and stopping the algorithm
  early when the prediction error begins to climb (indicating overfitting).
\end{enumerate} Psuedocode and Matlab code implementing this algorithm can be
found in the supplement.

In the boosting context, the property of refitting the ensemble weights in
every iteration is known as the \emph{totally corrective} property; LPBoost
[12] is a well-known example of a totally corrective boosting algorithm.  While
we derived EBP as a totally corrective variant of $L_2$Boost, one could also
view EBP as a generalization of the classical Lawson-Hanson (LH) algorithm [11]
for solving nonnegative least-squares problems. Given mild regularity
conditions and appropriate regularization, Elastic Basis Pursuit can be shown
to deterministically converge to the global optimum: we can bound the objective
function gap in the $m$th iteration by $C/\sqrt{m}$, where $C$ is an explicit
constant (see \ref{s:theory}).  To our knowledge, fixed iteration guarantees
are unavailable for all other methods of comparable generality for fitting a
mixture with an unknown number of components.

\subsection{Convergence Results}\label{s:theory}

\emph{(Detailed proofs can be found in the supplementary material.)}

For our convergence results to hold, we require an oracle function
$\tau: \mathbb{R}^{\tilde{n}} \to \Theta$ which satisfies

\begin{equation}\label{taucondition}
\left\langle \tilde{r}, \frac{\tilde{f}_{\tau(\tilde{r})}}{||\tilde{f}_{\tau(\tilde{r})}||}
\right\rangle \geq \alpha \rho(\tilde{r}) \text{, where }
\rho(\tilde{r}) = 
\sup_{\theta \in \Theta} \left\langle
\tilde{r}, \frac{\tilde{f}_\theta}{||\tilde{f}_\theta||}\right\rangle
\end{equation}
for some fixed $ 0 < \alpha <= 1$.  Our proofs can also be modified to apply
given a stochastic oracle that satisfies \eqref{taucondition} with fixed
probability $p > 0$ for every input $\tilde{r}$.  Recall that $\tilde{y}$
denotes the transformed input, $\tilde{f}_\theta$ the transformed kernel and
$\tilde{n}$ the dimensionality of $\tilde{y}$. We assume that the parameter
space $\Theta$ is compact and that $\tilde{f}_\theta$, the transformed kernel
function, is continuous in $\theta$.  Furthermore, we assume that either $L_1$
regularization is imposed, \emph{or} the kernels satisfy a positivity
condition, i.e. $\inf_{\theta \in \Theta} f_\theta(x_i) \geq 0$ for $i =
1,\hdots,n$. Proposition 1 states that these conditions imply the existence of
a maximally saturated model $(w^*,\boldsymbol{\theta}^*)$ of size $K^* \leq
\tilde{n}$ with residual $\tilde{r}^*$.

The existence of such a saturated model, in conjunction with existence of the
oracle $\tau$, enables us to state fixed-iteration guarantees on the precision
of EBP, which implies asymptotic convergence to the global optimum. To do so,
we first define the quantity $\rho^{(m)} = \rho(\tilde{r}^{(m)})$, see
\eqref{taucondition} above.  Proposition 2 uses the fact that the residuals
$\tilde{r}^{(m)}$ are orthogonal to $\tilde{F}^{(m)}$, thanks to the NNLS
fitting procedure in step 2.  This allows us to bound the objective function
gap in terms of $\rho^{(m)}$.  Proposition 3 uses properties of the oracle
$\tau$ to lower bound the progress per iteration in terms of $\rho^{(m)}$.

\noindent\textbf{Proposition 2}
\emph{
Assume the conditions of Proposition 1.
Take saturated model $w^*, \boldsymbol{\theta}^*$. Then defining}
\begin{equation}\label{Bstar}
B^* = 2 \sum_{i=1}^{K^*} w_i^* ||\tilde{f}_{\theta_i^*}||
\end{equation}
\emph{
the $m$th residual of the EBP
  algorithm $\tilde{r}^{(m)}$ can be bounded in size by
}
\[
||\tilde{r}^{(m)}||^2 \leq ||\tilde{r}^*||^2 + B^*\rho^{(m)}
\]

In particular, whenever $\rho$ converges to 0, the algorithm converges to the
global minimum.

\noindent\textbf{Proposition 3}
\emph{
Assume the conditions of Proposition 1. Then
\[
||\tilde{r}^{(m)}||^2 - ||\tilde{r}^{(m+1)}||^2 \geq (\alpha \rho^{(m)})^2
\]
for $\alpha$ defined above in \eqref{taucondition}. This implies that the sequence $||\tilde{r}^{(0)}||^2,\hdots$ is decreasing.
}

Combining Propositions 2 and 3 yields our main result for the non-asymptotic
convergence rate.

\noindent\textbf{Proposition 4}
\emph{Assume the conditions of Proposition 1.
Then for all $m > 0$,
\[
||\tilde{r}^{(m)}||^2 - ||\tilde{r}^*||^2 \leq 
\frac{B_{min}\sqrt{||\tilde{r}^{(0)}||^2 - ||\tilde{r}^*||^2||}}{\alpha}
\frac{1}{\sqrt{m}}
\]
where 
\[
B_{min} = \inf_{w^*,\boldsymbol{\theta}^*} B^*
\]
for $B^*$ defined in \eqref{Bstar}
}

Hence we have characterized the non-asymptotic convergence of EBP at rate
$\frac{1}{\sqrt{m}}$ with an explicit constant, which in turn implies
asymptotic convergence to the global minimum.

\section{DWI Results and Discussion}\label{s:results}

\begin{figure}[htbp]
\centering
\includegraphics[scale=0.3]{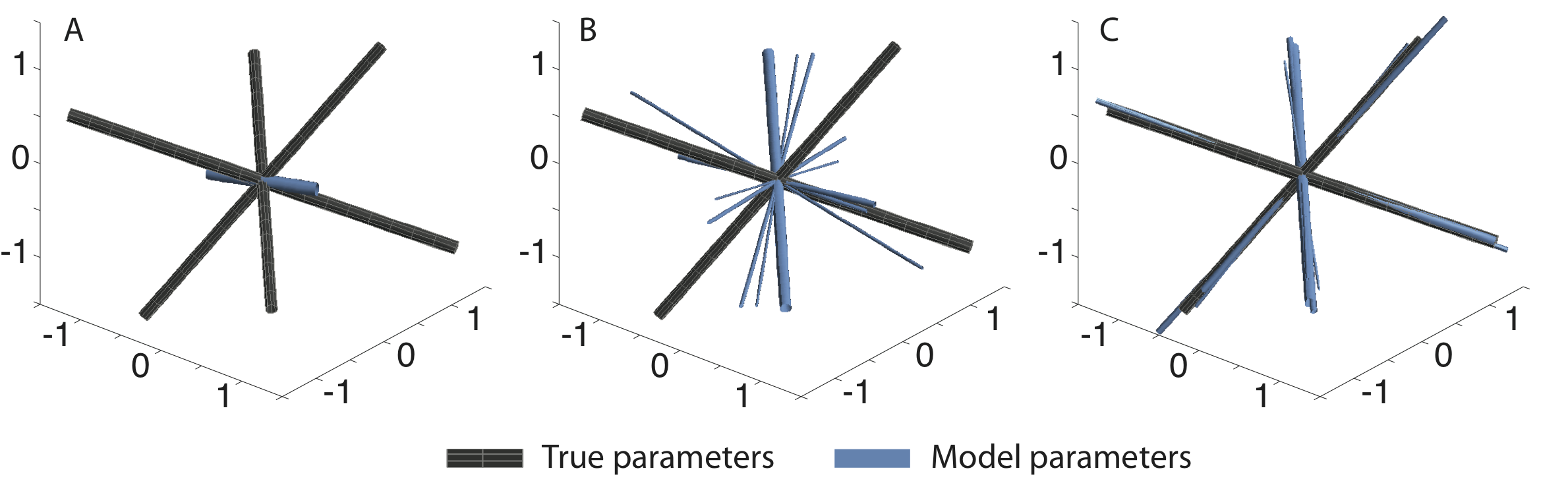}
\caption {DWI Simulation results. Ground truth entered into the simulation is a
  configuration of 3 crossing fascicles (A). DTI estimates a single primary
  diffusion direction that coincides with none of these directions (B). NNLS
  estimates an fODF with many, demonstrating the discretization error (see also Figure
  \ref{fig:illus1}). EBP estimates a much sparser solution with weights
  concentrated around the true peaks (D).}
\label{fig:results_sim}
\end{figure}

To demonstrate the utility of EBP in a real-world application, we used this
algorithm to fit mixture models of DWI. Different approaches are taken to
modeling the DWI signal. The classical Diffusion Tensor Imaging (DTI) model
[5], which is widely used in applications of DWI to neuroscience questions, is
not a mixture model. Instead, it assumes that diffusion in the voxel is well
approximated by a 3-dimensional Gaussian distribution. This distribution can be
parameterized as a rank-2 tensor, which is expressed as a 3 by 3
matrix. Because the DWI measurement has antipodal symmetry, the tensor matrix
is symmetric, and only 6 independent parameters need to be estimated to specify
it. DTI is accurate in many places in the white matter, but its accuracy is
lower in locations in which there are multiple crossing fascicles of nerve
fibers. In addition, it should not be used to generate estimates of
connectivity through these locations. This is because the peak of the fiber
orientation distribution function (fODF) estimated in this location using DTI
is not oriented towards the direction of any of the crossing fibers. Instead,
it is usually oriented towards an intermediate direction (Figure \ref{fig:results_sim}B). To
address these challenges, mixture models have been developed, that fit the
signal as a combination of contributions from fascicles crossing through these
locations. These models are more accurate in fitting the signal. Moreover,
their estimate of the fODF is useful for tracking the fascicles through the
white matter for estimates of connectivity. However, these estimation
techniques either use different variants of NNLS, with a discrete set of
candidate directions [2], or with a spherical harmonic basis set [1], or use
stochastic algorithms [9]. To overcome the problems inherent in these
techniques, we demonstrate here the benefits of using EBP to the estimation of
a mixture models of fascicles in DWI. We start by demonstrating the utility of
EBP in a simulation of a known configuration of crossing fascicles. Then, we
demonstrate the performance of the algorithm in DWI data.

The DWI measurements for a single voxel in the brain are $y_1,\hdots,y_n$ for
directions $x_1,\hdots,x_n$ on the three dimensional unit sphere, given by
\begin{equation}\label{dwi_convolution}
y_i = \sum_{k=1}^K w_k f_{D_k}(x_i) + \epsilon_i, \text{ where } f_D(x) = \exp[-bx^T D x],
\end{equation}
The kernel functions $f_D(x)$ each describe the effect of a single fascicle
traversing the measurement voxel on the diffusion signal, well described by the
Stejskal-Tanner equation [13]. Because of the non-negative nature of the MRI
signal, $\epsilon_i > 0$ is generated from a Rician distribution [14]. where
$b$ is a scalar quantity determined by the experimenter, and related to the
parameters of the measurement (the magnitude of diffusion sensitization applied
in the MRI instrument). $D$ is a positive definite quadratic form, which is
specified by the direction along which the fascicle represented by $f_D$
traverses the voxel and by additional parameters $\lambda_1$ and $\lambda_2$,
corresponding to the axial and radial diffusivity of the fascicle represented
by $f_D$. The oracle function $\tau$ is implemented by Newton-Raphson
with random restarts.
In each iteration of the algorithm, the parameters of $D$ (direction
and diffusivity) are found using the oracle function, $\tau(\tilde{r})$, using
gradient descent on $\tilde{r}$, the current residuals. In each iteration, the
set of $f_D$ is shrunk or expanded to best match the signal.

\begin{figure}[htbp]
\centering
\includegraphics[scale=0.7]{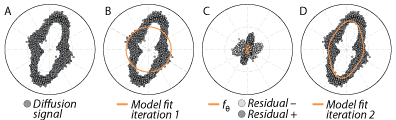}
\caption{ To demonstrate the steps of EBP, we examine data from 100 iterations
  of the DWI simulation. (A) A cross-section through the data. (B) In the first
  iteration, the algorithm finds the best single kernel to represent the data
  (solid line: average kernel). (C) The residuals from this fit (positive in
  dark gray, negative in light gray) are fed to the next step of the algorithm,
  which then finds a second kernel (solid line: average kernel). (D) The signal
  is fit using both of these kernels (which are the \emph{active set} at this
  point). The combination of these two kernels fits the data better than any of
  them separately, and they are both kept (solid line: average fit), but
  redundant kernels can also be discarded at this point (D).}
\label{fig:polar}
\end{figure}

\begin{figure}[htbp]
\centering
\includegraphics[scale=0.22]{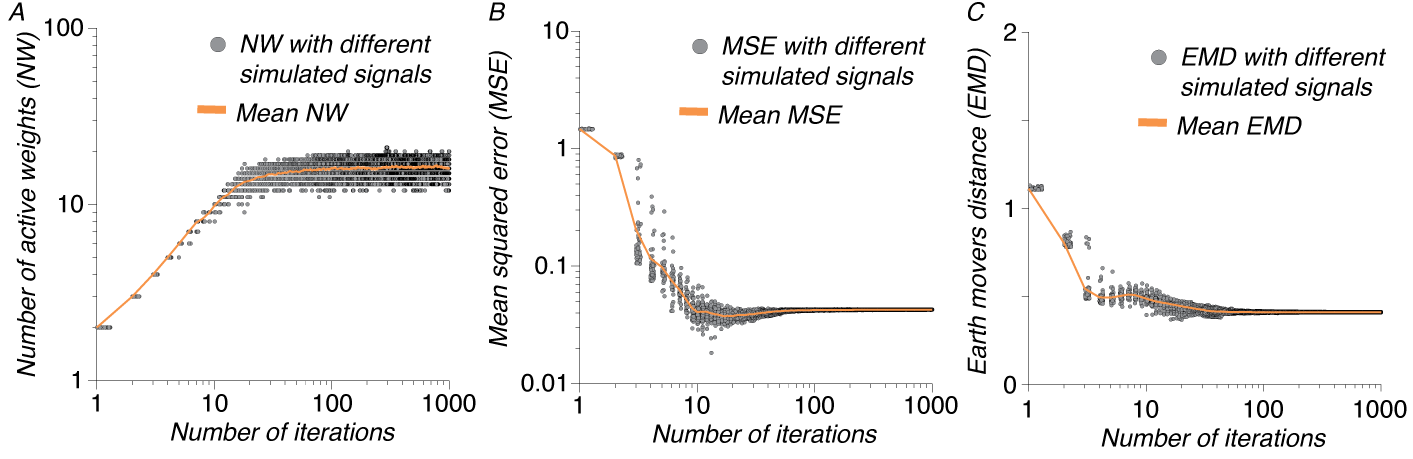}
\caption{The progress of EBP. In each plot, the abscissa denotes the number of
  iterations in the algorithm (in log scale). (A) The number of kernel
  functions in the active set grows as the algorithm progresses, and then
  plateaus. (B) Meanwhile, the mean square error (MSE) decreases to a minimum
  and then stabilizes. The algorithm would normally be terminated at this
  minimum. (C) This point also coincides with a minimum in the optimal
  bias-variance trade-off, as evidenced by the decrease in EMD towards this
  point.}
\label{fig:iterations}
\end{figure}

In a simulation with a complex configuration of fascicles, we demonstrate that
accurate recovery of the true fODF can be achieved.  In our simulation model,
we take $b=1000 s/mm^2$, and generate $v_1,v_2,v_3$ as uniformly distributed
vectors on the unit sphere and weights $w_1,w_2,w_3$ as i.i.d. uniformly
distributed on the interval $[0,1]$. Each $v_i$ is associated with a
$\lambda_{1,i}$ between 0.5 and 2, and setting $\lambda_{2,i}$ to 0.  We
consider the signal in 150 measurement vectors 
distributed on the unit sphere according to an electrostatic repulsion
algorithm. We partition the vectors into a training partition and a test
partition to minimize the maximum angular separation in each partition.  $\sigma^2 = 0.005$ we generate a signal 
 
We use cross-validation on the training set to fit NNLS with varying L1
regularization parameter $c$, using the regularization penalty function:
$\lambda P(w) =\lambda (c-||w||_1)^2$. We choose this form of penalty function
because we interpret the weights $w$ as comprising partial volumes in the
voxel; hence $c$ represents the total volume of the voxel weighted by the
isotropic component of the diffusion.  We fix the regularization penalty
parameter $\lambda=1$.  The estimated fODFs and predicted signals are obtained
by three algorithms: DTI, NNLS, and EBP. Each algorithm is applied to the
training set (75 directions), and error is estimated, relative to a prediction
on the test set (75 directions). The latter two methods (NNLS, EBP) use the
regularization parameters $\lambda = 1$ and the $c$ chosen by cross-validated
NNLS. Figure \ref{fig:polar} illustrates the first two iterations of EBP
applied to these simulated data.  The estimated fODF are compared to the true
fODF by the antipodally symmetrized Earth Mover's distance (EMD) [15] in each
iteration. Figure \ref{fig:iterations} demonstrates the progress of the
internal state of the EBP algorithm in many repetitions of the simulation. In
the simulation results (Figure \ref{fig:results_sim}), EBP clearly reaches a
more accurate solution than DTI, and a sparser solution than NNLS.

\begin{figure}[htbp]
\centering
\includegraphics[scale=0.5]{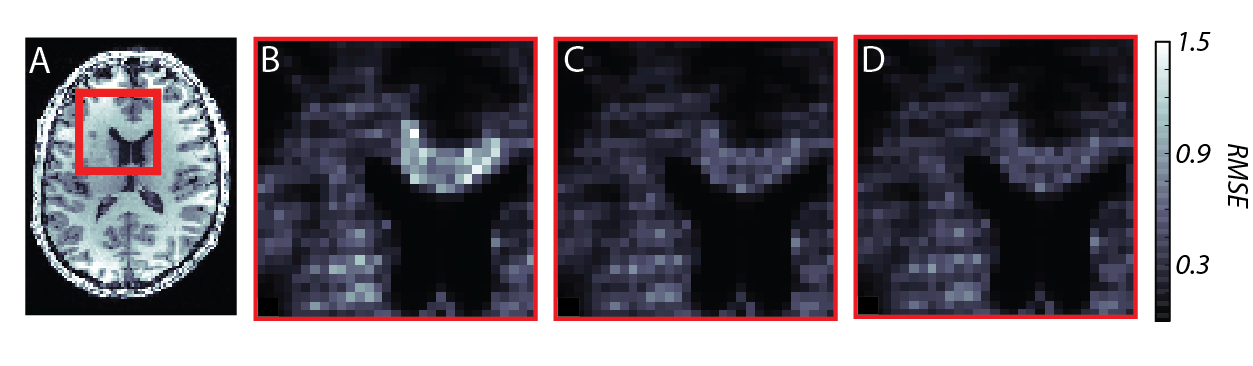}
\caption {DWI data from a region of interest (A, indicated by red frame) is
  analyzed and RMSE is displayed for DTI (B), NNLS(C) and EBP(D).}
\label{fig:results_data}
\end{figure}

The same procedure is used to fit the three models to DWI data, obtained at
2x2x2 $mm^3$, at a b-value of 4000 $s/mm^2$.  In the these data, the true fODF
is not known. Hence, only test prediction error can be obtained. We compare
RMSE of prediction error between the models in a region of interest (ROI) in
the brain containing parts of the corpus callosum, a large fiber bundle that
contains many fibers connecting the two hemispheres, as well as the centrum
semiovale, containing multiple crossing fibers (Figure \ref{fig:results_data}). NNLS
and EBP both have substantially reduced error, relative to DTI. 

\section{Conclusions}
We developed an algorithm to model multi-dimensional mixtures. This algorithm,
\emph{Elastic Basis Pursuit} (EBP), is a combination of principles from
boosting, and principles from the Lawson-Hanson \emph{active set} algorithm. It
fits the data by iteratively generating and testing the match of a set of
candidate kernels to the data. Kernels are added and removed from the set of
candidates as needed, using a totally corrective backfitting step, based on the
match of the entire set of kernels to the data at each step. We show that the
algorithm reaches the global optimum, with fixed iteration guarantees. Thus, it
can be practically applied to separate a multi-dimensional signal into a sum of
component signals.  For example, we demonstrate how this algorithm can be used to fit
diffusion-weighted MRI signals into nerve fiber fascicle components.

\subsubsection*{Acknowledgments}
The authors thank Brian Wandell and Eero Simoncelli for useful discussions. CZ
was supported through an NIH grant 1T32GM096982 to Robert Tibshirani and Chiara Sabatti,
AR was supported through NIH
fellowship F32-EY022294. FP was supported through NSF grant BCS1228397 to Brian
Wandell

\subsubsection*{References}

[1] Tournier J-D, Calamante F, Connelly A (2007). Robust determination of the
fibre orientation distribution in diffusion MRI: non-negativity constrained
super-resolved spherical deconvolution. {\it Neuroimage} 35:1459–72

[2] Dell’Acqua F, Rizzo G, Scifo P, Clarke RA, Scotti G, Fazio F (2007). A
model-based deconvolution approach to solve fiber crossing in
diffusion-weighted MR imaging. {\it IEEE Trans Biomed Eng} 54:462–72

[3] Ekanadham C, Tranchina D, and Simoncelli E. (2011). Recovery of
sparse translation-invariant signals with continuous basis pursuit.
{\it IEEE transactions on signal processing} (59):4735-4744.

[4] B{\"u}hlmann P, Yu B  (2003).  Boosting with the L2 loss: regression
and classification.  {\it JASA}, 98(462), 324-339.

[5] Basser,P. J., Mattiello, J. and Le-Bihan, D. (1994).  MR diffusion tensor
spectroscopy and imaging. {\it Biophysical Journal}, 66:259-267.

[6] Cand{\`e}s, E. J., and Fernandez‐Granda, C. (2013). Towards a Mathematical
Theory of Super‐resolution. \emph{Communications on Pure and Applied Mathematics.}

[7] Valiant, G., and Valiant, P. (2011, June). Estimating the unseen: an
n/log (n)-sample estimator for entropy and support size, shown optimal
via new CLTs. In \emph{Proceedings of the 43rd annual ACM symposium on
Theory of computing} (pp. 685-694). ACM.

[8] S{\'a}nchez-Bajo, F., and Cumbrera, F. L. (2000). Deconvolution of X-ray
diffraction profiles by using series expansion. {\it Journal of applied
crystallography}, 33(2), 259-266.

[9] Behrens TEJ, Berg HJ, Jbabdi S, Rushworth MFS, and Woolrich MW (2007).  Probabilistic
diffusion tractography with multiple fiber orientations: What can we
gain?  {\it NeuroImage} (34):144-45.

[10] Bro, R., and De Jong, S. (1997). A fast non-negativity-constrained least
squares algorithm. {\it Journal of chemometrics}, 11(5), 393-401.

[11] Lawson CL, and Hanson RJ. (1995). {\it Solving Least Squares
  Problems}.  SIAM.

[12] Demiriz, A., Bennett, K. P., and Shawe-Taylor, J. (2002). Linear
programming boosting via column generation. {\it Machine Learning}, 46(1-3), 225-254.

[13] Stejskal EO, and Tanner JE.  (1965). Spin diffusion measurements: Spin echoes
in the presence of a time-dependent gradient field.  {\it J Chem
  Phys}(42):288-92.

[14] Gudbjartsson, H., and Patz, S. (1995). The Rician distribution of noisy MR
data. {\it Magn Reson Med}. 34: 910–914.

[15] Rubner, Y.,   Tomasi, C.  Guibas, L.J.  (2000).  The earth mover's
distance as a metric for image retrieval.  {\it Intl J. Computer
  Vision}, 40(2), 99-121.

\section{Supplemental Material for Continuous Basis Pursuit for High
  Dimensional Mixtures}

\subsection{Continuous Basis Pursuit}

Continuous basis pursuit,
introduced by Ekanadham et al. [3],
can be viewed as an extension of nonnegative least squares where we
are given the liberty of perturbing
the points on the discretization grid $\vartheta_1,\hdots,\vartheta_p$
to adjusted versions $\tilde{\vartheta}_1,\hdots,\tilde{\vartheta}_p$ where the
perturbations are constrained to lie within Voronoi cells
$V_1,\hdots,V_p$ generated by $\vartheta_1,\hdots,\vartheta_p$.
The idea of CBP is to linearly approximate the resulting kernel functions
$f_{\tilde{\vartheta}_i}(x)$.
In particular, in first-order CBP (FOCBP), one uses the approximation
\[
f_{\tilde{\vartheta}_i}(x) \approx \tilde{f}_{\tilde{\vartheta}_i}(x)
= f_{\vartheta_i}(x) + \sum_{d=1}^D (\tilde{\vartheta}_i -
\vartheta_i)_d \frac{\partial f_\theta(x)}{\partial (\theta)_d}\bigg|_{\vartheta_i}
\]
where $D$ is the dimensionality of the parameter space.
Defining $X_{i,0} =
(f_{\vartheta_i}(x_1),\hdots,f_{\vartheta_i}(x_n))$ and 
\[X_{i,d} = \left(\frac{\partial f_\theta(x_1)}{\partial
  (\theta)_d}\bigg|_{\vartheta_i},\hdots,
\frac{\partial f_\theta(x_n)}{\partial
  (\theta)_d}\bigg|_{\vartheta_i}\right)\]
for $d = 1,\hdots,D$, and convex constraint sets
\[
C_i = \{(x,z) \in \mathbb{R}\times\mathbb{R}^d: \vartheta_i +
\frac{z}{x} \in V_p\}
\]
one writes the FOCBP objective function as
\begin{equation}\label{focbp}
\text{minimize }_{\beta} ||y-X\beta||^2 + \lambda P_\theta(\beta_{\cdot,0})
\end{equation}
subject to
\begin{align*}
\beta_{i,0} &\geq 0\\
(\beta_{i,0},\beta_{i,1},\hdots,\beta_{i,d}) &\in C_i 
\end{align*}
yielding estimates $\hat{K} = \sum_{i=1}^p I(\beta_{i,0} > 0)$,
\[
\hat{\theta} = \left(\vartheta_i + \sum_{d=1}^D
\frac{\beta_{i,d}}{\beta_{i,0}}e_d : \beta_{i,0} > 0\right)
\]
\[
\hat{w} = \left(\beta_{i,0}: \beta_{i,0} > 0\right)
\]
where $e_d$ is the $d$th standard basis vector.

Ekanadham et al. suggested using solvers for semidefinite programs
(SDP) to solve instances of CBP problems, like the objective function above.
However, we found that FOCBP can be transformed into a
nonnegative least squares problem, generally resulting in speedups and
improvements in stability.

The key observation is that any pertubed parameter $\vartheta_i \in
V_i$ can be represented as a positive linear combination of the finite
set of vertices of $V_i$, $v_{i,1},\hdots,v_{i,m_i}$.
Yet, this implies that the corresponding approximated kernel function
$\tilde{f}_{\vartheta_i}$ can also be represented as a positive linear
combination of $\tilde{f}_{v_{i,1}},\hdots,\tilde{f}_{v_{i,m_i}}$.

Hence defining $Z_{i,1},\hdots,Z_{i,m_i}$ by
\[
Z_{i,j} = \left(
\sum_{d=1}^D (v_{i,j}-\vartheta_i)_d\frac{\partial
  f_\theta(x_1)}{\partial (\theta)_d}\bigg|_{\vartheta_i}
,\hdots,
\sum_{d=1}^D (v_{i,j}-\vartheta_i)_d\frac{\partial
  f_\theta(x_1)}{\partial (\theta)_d}\bigg|_{\vartheta_i}
\right)
\]
we can obtain the equivalent problem
\[
\text{minimize }_{\gamma > 0} ||y-Z\gamma||^2 + \lambda P(\gamma)
\]
which yields identical estimates to the original approach
\eqref{focbp}, via 
\[\hat{K} = \sum_{i=1}^p I\left(\sum_{j=1}^{m_i} \gamma_{i,j} > 0\right) \]
\[
\hat{w} = \left(\sum_{j=1}^{m_i}  \gamma_{i,m_i} : \sum_{j=1}^{m_i} \gamma_{i,j} > 0\right)
\]
\[
\hat{\theta} = \left(
\frac{\sum_{j=1}^{m_i} \gamma_{i,j}v_{i,j}}{\sum_{j=1}^{m_i}
  \gamma_{i,j}} :\sum_{j=1}^{m_i} \gamma_{i,j} > 0
\right)
\]

\subsection{The Lawson-Hanson algorithm for positive mixture problems}\label{s:lh}

Before discussing our proposed method, true continuous basis pursuit,
we discuss the unique properties of the Lawson-Hanson algorithm
[11] for solving nonnegative least
squares problems of the form
\begin{equation}\label{LHprob}
\text{minimize}_\beta ||y-X\beta||^2 \text{ subject to }\beta \geq 0
\end{equation}
where $X$ is a $n \times p$ matrix,
in the special case of $X$ with nonnegative entries.

The algorithm begins with an active set $S$ initialized to the null
set and estimate $\beta$ intialized to 0, and uses a tolerance
$\epsilon > 0$.
Letting $X_S$ represent the columns of $X$ corresponding to the
indices included in $S$ and $\beta_S$ be the entries of $\beta$
corresponding to the indices of $S$.
The LH algorithm is as follows [Charles will summarize]

\noindent\textbf{Initialization}
\begin{enumerate}
\item Initialize set $S$ of indices to the empty set.
\item Intiialize $\beta$ to be a $p \times 1$ vector of zeroes
\item Initialize $w = X^T(y-X\beta)$ 
\item Run main loop
\item Return $\beta$, the solution to the least-squares problem \eqref{LHprob}
\end{enumerate}

\noindent\textbf{Main Loop}
\begin{enumerate}
\item \emph{While} $\max(w) > \epsilon$:
\item Letting $j$ be the smallest index such that $w_j = \max(w)$, set
  $S \leftarrow S \cup \{j\}$
\item Let $s$ be a $p \times 1$ vector of zeros.
\item Set $s_S \leftarrow (X_s^T X_S)^{-1} X_S^T y$
\item Begin \textbf{inner loop}.
\item Set $\beta \leftarrow s$.
\item Set $w \leftarrow X^T(y-X\beta)$
\item \emph{End while}
\end{enumerate}

\noindent\textbf{Inner Loop}
\begin{enumerate}
\item \emph{While} $\max(s) \geq 0$:
\item Let $I$ be the set of indices $i$ where $s_i < \beta_i$.
\item Let $\alpha = \min_{i \in I} \beta_i/(\beta_i-s_i)$
\item Set $\beta \leftarrow \beta + \alpha(s-\beta)$
\item Set $S \leftarrow \{i: \beta_i > 0\}$.
\item \emph{End while}
\end{enumerate}

Since the LH algorithm was proposed in 1974, a number of improvements
have been proposed for solving large-scale nonnegative least squares
problem.  Efron's least-angle procedure is especially suitable for
solving the lasso-regularized NNLS problem $\min_{\beta \geq 0} ||y-X\beta||^2
+ \lambda||\beta||_1$ but can also be applied to the original NNLS
problem.
Kim, Sra and Dhillon proposed an interior-point based method
for solving NNLS problems using conjugate gradients.  Potluru
propose using coordinate descent to solve NNLS. The FISTA
algorithm of Beck can also be modified to solve NNLS.

But in the special case of positive $X$ and $p >> n$, one can see both
theoretically and empirically that the original LH algorithm far
outperforms these more recent competing methods.

Firstly, the $\beta$ vector remains sparse in every iteration of the
LH algorithm, even for noisy data.  This means that the LH algorithm
gains a substantial advantage over coordinate descent methods by
computing the true least-sqaures solution for the current active set.

Secondly, the nature of the basis set renders
gradient-descent based approaches, like the Kim Sra Dhillon algorithm, much
less effective. Due to the high degree of collinearity in the basis
set, the function has high curvature in the direction of the gradient,
which often reduces the maximum step size at each iteration to below
working precision.

Thirdly, the nonnegativity constraints combined with high
dimensionality pose a challenge to methods like FISTA, which rely on
log barrier functions to enforce the nonnegativity constraint.

Fourthly, the geometry of the basis set, which resembles a high-dimensional
connected, curved surface with a spike at $(1,\hdots,1)$, poses
special difficulties for
Efron's LARS algorithm, which aggresively adds variables to the active
set as it continuously adjusts the coefficients of the solution
vector.  The LARS algorithm is hampered by the frequency at which the
active set must change along the solution path.  On the other hand,
since the LARS algorithm recovers the entire L1 regularized solution
path, it may still be useful for tuning the L1 regularization parameter.

\subsection{Proofs}

Recall that we define an oracle $\tau: \mathbb{R}^n \to \Theta$ via the
property that
\begin{equation}\label{taucondition}
\langle r, \vec{f}_{\tau(r)} \rangle \geq \alpha \max_\Theta \frac{\langle
r, \vec{f}_\theta \rangle}{||\vec{f}_\theta||}
\end{equation}
for some fixed $\alpha > 0$.

\noindent\textbf{Proposition}.
\emph{For any positive integer $K \geq n$, and for any
  $w \in \mathbb{R}_+^K,\boldsymbol{\theta} \in \Theta^K$,
there exists $\tilde{w}, \tilde{\boldsymbol{\theta}} \in \Theta^n$
such that}
\[
L(\tilde{w},\tilde{\boldsymbol{\theta}}) \leq L(w,\boldsymbol{\theta})
\]
\emph{for $L$ defined in \eqref{lsp0obj}.}

\noindent\textbf{Proof}.
Form the matrix $\vec{F} = (\vec{f}_{\theta_1},\hdots,
\vec{f}_{\theta_K})$.
Then
\[
L(\beta,\boldsymbol{\theta}) = ||y-\vec{F}\beta||^2
\]
for any $\beta \in [0,\infty)^K$.
But if we minimize $||y-X\beta||^2$ over $\beta$ nonnegative,
we can find a solution $\beta^*$ with $n$ or fewer nonzero entries, as
proved in Lawson and Hanson [7].
Taking $\tilde{w}$ to be the nonnegative entries of $\beta^*$ and taking
$\tilde{\boldsymbol{\theta}}$ to be the corresponding parameters
$\theta$,
we have $L(\tilde{w},\tilde{\boldsymbol{\theta}}) = ||y-X\beta^*||^2
\leq L(\tilde{w},\boldsymbol{\theta})$. $\Box$

For the Lemma, we take $\Theta$ to be a compact set in $\mathbb{R}^D$ and we require
that $f_\theta(x)$ be continuous with respect to $\theta$ for any
fixed $x$.

\noindent\textbf{Lemma.}
\emph{Under the conditions stated above, there exists a nonnegative
  integer $K^* \leq n$ and $w^* = (w^*_1,\hdots,w^*_{K*})$ and
  $\boldsymbol{\theta}^* = (\theta^*_1,\hdots,\theta^*_{K^*})$ such that }
\begin{equation}\label{minimizer}
\left\| y - \sum_{i=1}^{K^*} w^*_i \vec{f}_{\theta^*_i} \right\|^2 = 
\inf_{w,\theta, K\leq n} 
\left\| y - \sum_{i=1}^{K} w_i \vec{f}_{\theta_i} \right\|^2
\end{equation}

\noindent\textbf{Proof.}
Since $\Theta$ is compact, so is $[0,\infty)^n \times \Theta^n$.
Also the space $\{\vec{f}_\theta \in \mathbb{R}^n : \theta \in
\Theta\}$ is compact.
And by the continuity of $f$,
if we define $L: [0,\infty)^n \times \Theta^n \to \mathbb{R}$ by

then $L$ is continuous.
Since the squared norm of any vector is nonnegative, we know that
$\inf_{w,\boldsymbol{\theta}} L \geq 0$.
By the compactness of $[0,\infty)^n \times \Theta^n$, there exists
$w,\theta$ such that $L(w,\boldsymbol{\theta}) = \inf_{w,\boldsymbol{\theta}} L(w,\boldsymbol{\theta})$.
Take $K^* = \sum_{i=1}^n I(w_i \neq 0)$ and take $w^*$ to be the
sequence of nonnegative entries of $w$, and $\boldsymbol{\theta}^*$ to be the
sequence of nonnegative entries of $\boldsymbol{\theta}$ to complete the
proof. $\Box$

\noindent\textbf{Proposition.}
\emph{Suppose there exists $w^*, \boldsymbol{\theta}^*$ satisfying
  \eqref{minimizer}.  Then for any oracle $\tau$ satisfying
  condition \eqref{taucondition} there exists $C \in \mathbb{R}$ and $M \in
  \mathbb{N}$ such that for all iterations $m > M$ of the LH algorithm, we have
}
\[
||r^{(m)}||^2 < C/\sqrt{m}
\]

\noindent\textbf{Proof.}
For $m=1,2,\hdots$ define 
\[
\rho^{(m)} = \max_{\theta \in Theta} \langle r^{(m)}, \frac{\vec{f}_\theta}{||\vec{f}_\theta||}\rangle
\]
First we show that $\rho^{(m)}$ produces an upper bound on
$L(w^{(m)},\boldsymbol{\theta}^{(m)}) - L(w*,\boldsymbol{\theta}^*)$.
Define
\[
h^{(m)}(x,z) = \left\| r^{(m)} - \sum_{i=1}^{K^{(m)}} x_i
  \vec{f}_{\theta_i^{(m)}} - \sum_{i=1}^{K*} z_i \vec{f}_{\theta_i^*} \right\|
\]
Note that $h$ is jointly convex in $(x,z)$, and verify that $h^{(m)}(0,0) = L(w^{(m)},\boldsymbol{\theta}^{(m)})$ and
$h^{(m)}(-w^{(m)},w^*) = L(w*,\boldsymbol{\theta}^*)$.
Further note that
\[
\frac{\partial h^{(m)}}{x_i} = 0
\]
due to the fact that the residual $r^{(m)}$ is orthogonal to the
columns of $\vec{F}^{(m)}$ (see [7]).
Meanwhile, note that $\langle r^{(m)},
\vec{f}_i^* \rangle < \rho ||\vec{f}_i^*||$, which implies
\[
\frac{\partial h^{(m)}}{z_i} \geq -2\sqrt{K^*}\rho
\]
Now due to the convexity of $h$, we have
\begin{align}
L(w^{(m)},\boldsymbol{\theta}^{(m)}) - L(w*,\boldsymbol{\theta}^*) &=
h(0,0)-h(-w^{(m)},w^*) \leq |\langle-w^{(m)}, \nabla_x h(0,0) \rangle
+ \langle w^*, \nabla_z h(0,0)\rangle| 
\\& \leq 2B^* \sqrt{K^*}\rho \label{gapbound}
\end{align}
where
\[
B^* = \sqrt{\sum_{i=1}^{K^*} (w^*_i ||\vec{f}^*_i||)^2}
\]
The next major step is to see that
\begin{align}
||r^{(m+1)}||^2 &= \min_{\beta > 0}||y - \vec{F}^{(m)}\beta||^2
\\&\leq \left\|y - \vec{F}^{(m-1)} \beta^{(m-1)} - \vec{f}_{\vartheta^{(m+1)}_1}\frac{\langle \vartheta^{(m+1)}, r^{(m)}
  \rangle}{||\vec{f}_{\vartheta^{(m+1)}}||||r^{(m)}||}\right\|^2 \label{sd1}
\\&= \left\|r^{(m)} - \vec{f}_{\vartheta^{(m+1)}_1}\frac{\langle \vartheta^{(m+1)}, r^{(m)}
  \rangle}{||\vec{f}_{\vartheta^{(m+1)}}||||r^{(m)}||}\right\|^2
\\& = ||r^{(m)}||^2 - \frac{\langle \vartheta^{(m+1)}, r^{(m)}
  \rangle}{||\vec{f}_{\vartheta^{(m+1)}}||} \label{sd2}
\\& \leq ||r^{(m)}||^2 - \left(\frac{\alpha
    \rho^{(m)}}{||r^{(m)}||}\right)^2 \label{sd3}
\\& \leq ||r^{(m)}||^2 - \left(\frac{\alpha
    \rho^{(m)}}{||y||}\right)^2 \label{sd4}
\end{align}
which implies
\begin{equation}\label{stepdecrease}
||r^{(m)}||^2 - ||r^{(m+1)}||^2 > \left(\frac{\alpha\rho^{(m)}}{||y||}\right)^2
\end{equation}
Here, \eqref{sd1} follows from the fact that the columns of $\vec{F}^{(m+1)}$ 
include $\vec{f}_{\vartheta^{(m+1)}_1}$ by also all of the columns of
$\vec{F}^{(m)}$ for which $\beta^{(m)}$ is nonzero.
Next, \eqref{sd2} is obtained by an application of the Pythagorean
theorem, and \eqref{sd3} by applying the definitions of $\rho^{(m)}$
and the condition \eqref{taucondition} on $\tau$.
Finally, \eqref{sd4} follows from observing that $||r^{(m)}||$ is
nondecreasing in $m$, hence $||r^{(m)}|| \leq ||y||$.

From this result, we obtain
\begin{align*}
||y||^2
&= \sum_{m=0}^\infty ||r^{m}||^2 - ||r^{m-1}||^2 
\\&= \sum_{m=0}^\infty \frac{\alpha^2 (\rho^{(m)})^2}{2\sqrt{K^*}B||y||^2}
\end{align*}

But since $||y||^2 < \infty$, this implies that $\sum_{m=0}^\infty
(\rho^{(m)})^2$ is convergent.
Hence, there exists a constant $C_0$, $\epsilon > 0$ and $M \in \mathbb{N}$ such that
for all $m  > M$,
\[
(\rho^{(m)})^2 \leq \frac{C_0}{m^{1+\epsilon}}
\]
Since we bounded the objective function gap in terms of $\rho^{(m)}$
in \eqref{gapbound}, this yields the desired result.


\end{document}